\documentclass[sigconf]{acmart}
\usepackage[utf8]{inputenc}
\usepackage[T1]{fontenc}
\usepackage{microtype}
\usepackage{mfirstuc}
\usepackage{subfig}
\usepackage{cleveref}

\newcommand{\workname}{Latent Diffusion}
\newcommand{\Workname}{Latent Diffusion}
\AtBeginDocument{%
  }

\setcopyright{acmlicensed}
\copyrightyear{2025}
\acmYear{2025}
\acmDOI{XXXXXXX.XXXXXXX}

\acmConference[]{}{}{}
\acmISBN{978-1-4503-XXXX-X/18/06}

\acmSubmissionID{artpl137s1}

\begin{document}

\title{Latent Diffusion : Multi-Dimension Stable Diffusion Latent Space Explorer}

\author{Zhihua Zhong}
\email{zzhong839@connect.hkust-gz.edu.cn}
\orcid{0009-0002-9979-4770}
\affiliation{%
  \institution{The Hong Kong University of Science and Technology (Guangzhou)}
  \country{China}
}

\author{Xuanyang HUANG}
\authornote{Corresponding author.}
\email{xhuang383@connect.hkust-gz.edu.cn}
\affiliation{%
  \institution{The Hong Kong University of Science and Technology (Guangzhou)}
  \country{China}
}

\renewcommand{\shortauthors}{Zhong et al.}

\begin{abstract}
Latent space is one of the key concepts in generative AI, offering powerful means for creative exploration through vector manipulation. However, diffusion models like Stable Diffusion lack the intuitive latent vector control found in GANs, limiting their flexibility for artistic expression. This paper introduces \workname, a framework for integrating customizable latent space operations into the diffusion process. By enabling direct manipulation of conceptual and spatial representations, this approach expands creative possibilities in generative art. We demonstrate the potential of this framework through two artworks, \textit{Infinitepedia} and \textit{Latent Motion}, highlighting its use in conceptual blending and dynamic motion generation. Our findings reveal latent space structures with semantic and meaningless regions, offering insights into the geometry of diffusion models and paving the way for further explorations of latent space.
\end{abstract}


\begin{teaserfigure}
  \includegraphics[width=\textwidth]{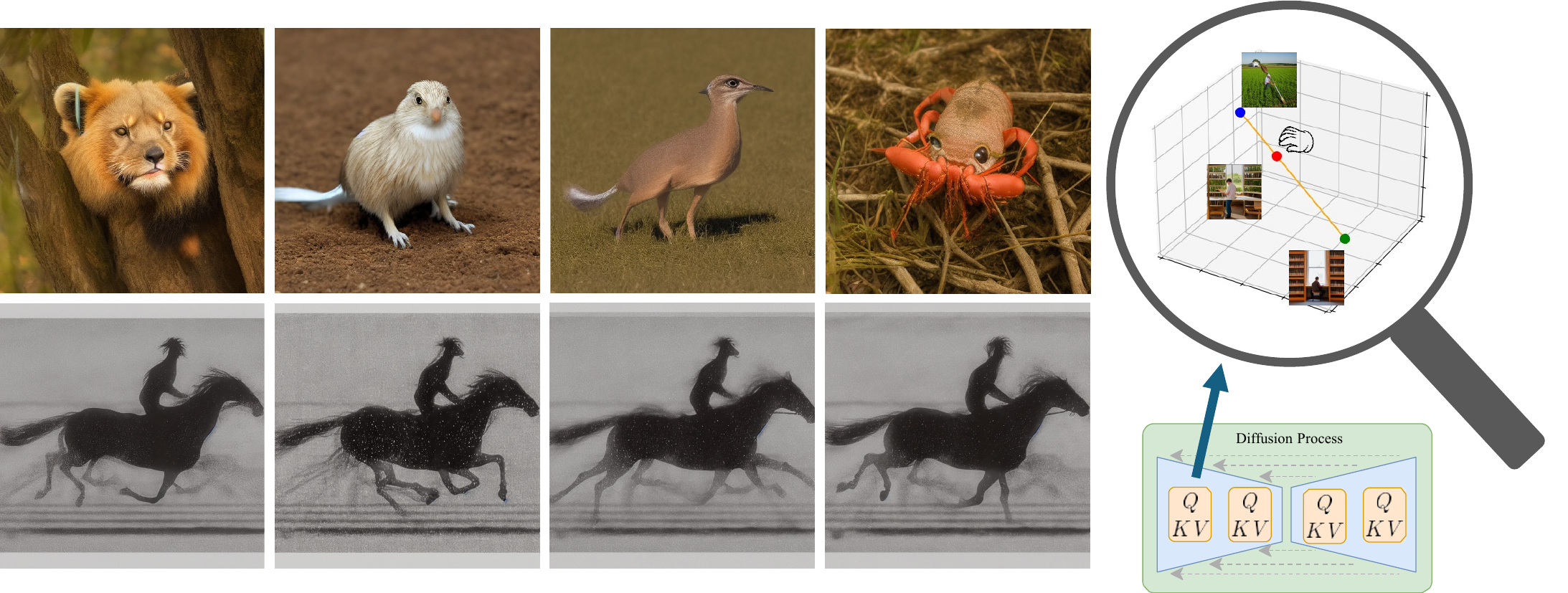}
  \caption{Latent Diffusion is a flexible framework for manipulating latent space vectors by intervening in the diffusion process. The left panel showcases results from artistic projects created with this tool, highlighting outputs generated through concept-based and shape-based explorations. The right panel illustrates latent space exploration, depicting vector manipulations enabled by custom operators. This framework empowers artists with precise and dynamic control over the generative process in Stable Diffusion.}
  \label{fig:teaser}
\end{teaserfigure}


\maketitle

\section{Introduction}
Latent space has emerged as a foundational concept in generative AI, enabling deep models to represent abstract, high-dimensional features that guide the synthesis of images, sound, and text. In the context of generative adversarial networks (GANs)\cite{goodfellow2014generative}, latent space manipulation allows artists and researchers to explore creative transformations by adjusting vectors that correspond to meaningful semantic changes. However, diffusion models, such as Stable Diffusion\cite{rombach2022high}, offer a fundamentally different architecture. Rather than directly exposing a latent vector for manipulation, they rely on iterative denoising guided by text-based prompts, making direct latent space exploration more challenging. This limits the creative flexibility available to users seeking fine-grained control over the generated content.

To address this gap, we introduce \workname, a novel tool that extends Stable Diffusion with direct latent space operations. We introduce two types of custom vector manipulations: \textbf{Query-wise Concept Latent Operation} and \textbf{Conditioning Vector Shape Latent Operation}. These operations provide precise control over conceptual and spatial representations, respectively, enabling users to purposefully explore the latent space along the dimensions of concept and shape and broadening the creative possibilities in diffusion-based generative art.

Our contributions can be summarized as follows:

\begin{itemize}
    \item Proposes a novel tool for introducing manipulable latent space vectors in Stable Diffusion, providing new possibilities for direct latent space exploration.

    \item Develop a flexible tool that enables manipulation of conceptual and spatial representations through customizable operators.
    
    \item Two artistic projects, \textit{Infinitepedia} and \textit{Latent Motion}, demonstrate the practical applications of this tool, highlighting the advantages of our work in terms of semantic robustness and reduced ambiguity inside latent space.

    \item The paper identifies the phenomenon of "latent deserts" within diffusion models and discusses potential geometric operations for future exploration.
\end{itemize}
\section{Background}

\subsection{Visual Indeterminacy in Art}

Visual Indeterminacy describes a perceptual phenomenon where observers can identify individual elements within an image but cannot integrate them into a coherent whole\cite{pepperell2006seeing}. Unlike Pareidolia, where people recognize familiar patterns in random stimuli, visual indeterminacy involves a sustained state of perceptual ambiguity despite the presence of identifiable features.

In AI art, this phenomenon takes on new significance through the lens of latent space. Hertzmann (2020) posits that the latent space in generative models like GANs serves as a creative domain rather than merely a mathematical construct \cite{hertzmann2020visual}. This continuous feature space allows artists to explore the boundary between the known and unknown, creating transitional states between different visual concepts. 

This approach highlights three key aspects of AI artistic creation: the ability to generate images that exist between recognition and ambiguity, the systematic exploration of novel visual forms through latent space continuity, and the strategic use of technical indeterminacy as a creative tool. The result is artwork that occupies a position between familiarity and uncanny.

\subsection{Latent Space Exploration in Generative Models}


In AI art, artists explore and manipulate complex high-dimensional representations within latent spaces of artificial intelligence systems. Numerous artistic works and projects have advanced the interpretative approaches to latent spaces and their transformation into visual outputs.

One of the earliest explorations of latent space is Mario Klingemann’s Memories of Passersby I (2018)\cite{klingemann2018memories}, an autonomous installation that continuously generates portraits of non-existent people using a GAN model. This work demonstrates how latent space can be traversed to interpolate facial features, blending memory and imagination into dynamic representations of human identity.





Artists explore visual possibilities within the latent spaces of GANs through diverse approaches, ranging from architectural data to historical portraits, from botanical patterns to biological forms. Through latent space manipulation, artists seek to discover fictional representations. Refik Anadol's "Machine Hallucinations" \cite{anadol2019machine} transforms the latent space of a neural network, trained on millions of architectural images, into immersive environments, presenting a spatial metaphor for machine perception and collective memory. Sofia Crespo's "Neural Zoo" \cite{crespo2018neural} employs AI-generated imagery inspired by biological patterns, exploring the creative boundaries of latent space through computational synthesis to envision hybrid organisms that do not exist in reality. Helena Sarin's "Leaves of Manifold"\cite{sarin2021leaves} constructs complex hybrid compositions by inputting personal sketches and photographs into GANs and manipulating latent vectors. In these works, latent space becomes a domain for artists to explore and reimagine real-world objects, architecture, and visual content.

Unlike the latent space exploration in the GAN era, artists and technologists are now exploring new methods to investigate the latent space within diffusion models. Kiss/Crash (2023) \cite{cole2023kiss} by Cole and Grierson exemplifies this through innovative video translation approaches in exploring the diffusion models' latent space. Departing from traditional diffusion models' denoising process, this work develops a series of techniques for controlled temporal transitions: the frame-blending method creates coherent trajectories between generated states, strength-scheduling enables gradual transitions from source images to target prompts, while prompt-interpolation achieves smooth latent space traversal through semantic guidance.


These artworks demonstrate how latent space functions as a dynamic field for creative and artistic manipulation, facilitating both conceptual exploration and technical innovation within the AI art community, as artists investigate the possibilities inherent in latent spaces.
However, these approaches face certain technical limitations. While GAN-based latent space vectors often have limited semantic interpretability, diffusion models like Stable Diffusion rely primarily on prompt-guided generation, making direct and precise latent space manipulation more challenging.

\subsection{Challenge in Stable Diffusion Latent Space Exploration}

The primary technical challenge in adding manipulable latent space vectors to Stable Diffusion, akin to those in GANs, lies in determining which latent vectors to manipulate. GANs inherently provide a manipulable latent vector: a noise vector designed to generate different outputs. Over time, this noise vector has become a representation within the network's latent space. This is partly because GANs rely exclusively on the noise vector as input, and partly due to the relatively shallow depth of GAN architectures, compared to Stable Diffusion, which primarily perform a form of “translation” rather than complex generation. Subsequent work, such as GAN-Control\cite{shoshan2021gan} and InfoGAN\cite{chen2016infogan}, explicitly treats this noise vector as a semantic representation, refining its parameters to allow more purposeful manipulation of the generated outputs.

In contrast, Stable Diffusion’s generation process is continuously guided by prompt embeddings and unfolds through multiple iterations. The initial noise input in Stable Diffusion acts merely as a random seed (unless intentionally under-iterated for artistic choice). Therefore, identifying suitable vectors for latent space exploration requires deliberate design choices.

In a neural network, every layer's output is a latent representation, produced by preceding layers and interpreted by subsequent ones. However, not all feature vectors are appropriate for latent space manipulation. Some vectors represent foundational stages of computation, and modifying them indiscriminately can cause the entire process to fail or result in noises\cite{szegedy2013intriguing}. Other vectors, though manipulable without causing network failure, may lack semantic relevance or interpretability.

Additionally, robustness and desirable mathematical properties are critical. Effective latent space manipulation should maintain exploratory potential even after complex vector operations, ensuring that modified vectors yield meaningful and interpretable results rather than collapsing into artifacts or random patterns.

\section{Methods}

\Workname introduces two types of latent space operations based on Stable Diffusion. The \workname accepts a latent space vector operator as input and modifies the image generation process within the network. This enables a custom processing of latent space information related to Stable Diffusion’s understanding of concepts or shapes. By doing so, the tool expands the possibilities and operational space for artists to explore latent spaces in Stable Diffusion, fostering creative experimentation and innovation.

The first operation, \textbf{Query-wise Concept Latent Operation}, manipulates latent space related to the AI's understanding of object concepts, allowing for perturbations in its conceptual cognition. The second, \textbf{Conditioning Vector Shape Latent Operation}, focuses on latent vectors associated with object shape and spatial information, enabling exploration of the AI's spatial perception. The following subsections describe these operations in detail.

\subsection{Latent Operator}
\begin{figure}
    \centering
    \includegraphics[width=0.95\linewidth]{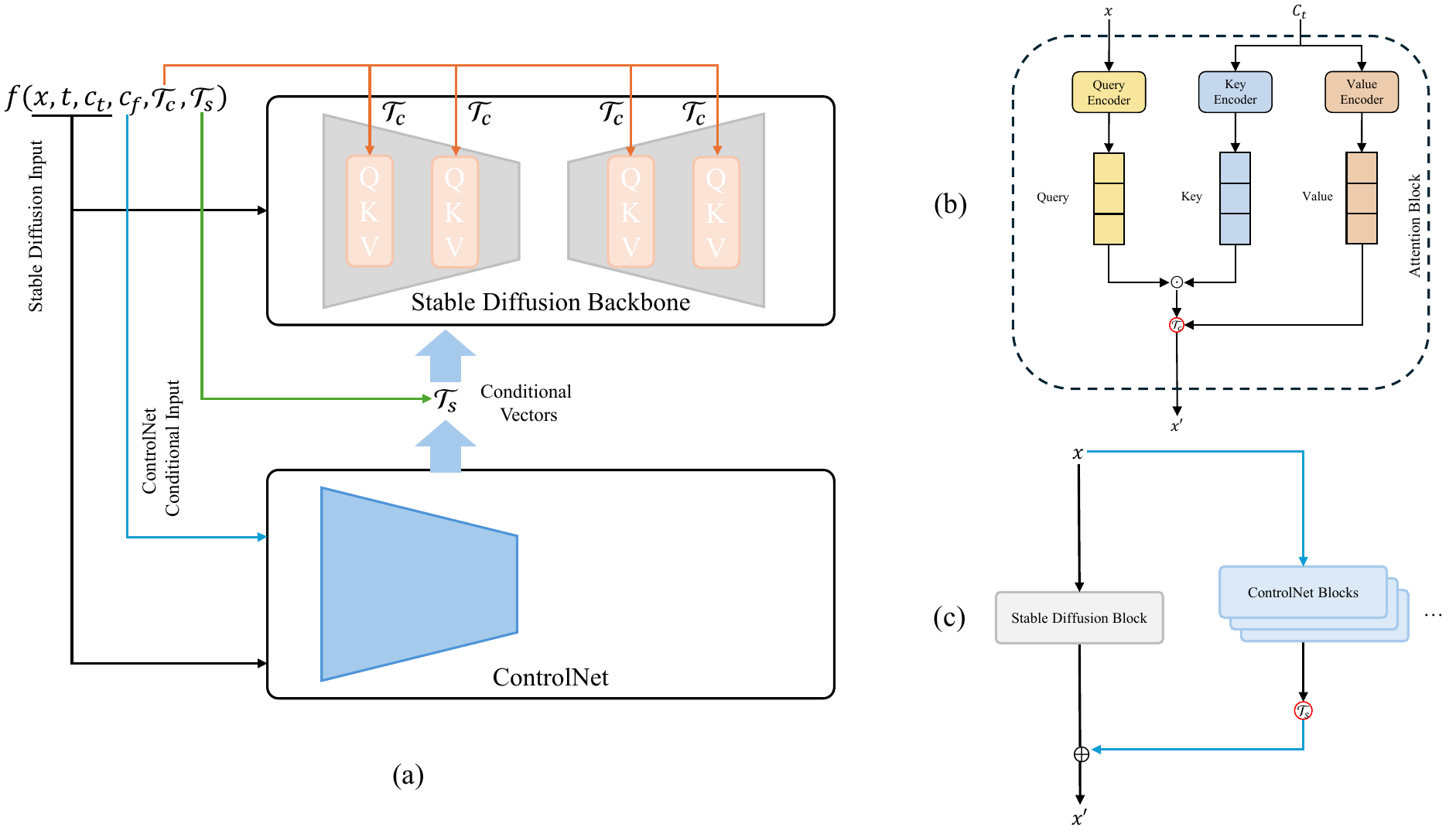}
    \caption{(a) is An overview of \workname architecture. (b) and (c) are details of \textbf{Query-wise Concept Latent Operation} and \textbf{Conditioning Vector Shape Latent Operation} respectively.}
    \label{fig:pipeline}
\end{figure}
\Workname extends the original Stable Diffusion pipeline by introducing two additional operators, $\mathcal{T}_c$ and $\mathcal{T}_s$, designed to control the neural network’s understanding of concepts and shapes. These operators are integrated into the Stable Diffusion workflow as \cref{fig:pipeline}(a) shows, where corresponding latent space vectors are fed into them. The operators' outputs are then passed as new latent space vectors to the next stage of the pipeline. If no operators are defined, the latent space vectors are forwarded unchanged, and the network’s understanding of the relevant information defaults to that of standard Stable Diffusion.

The input operators can be arbitrary vector operations, taking one or more latent space vectors as input and producing a single latent space vector as output. In our artistic works, \textit{Infinitepedia} and \textit{Latent Motion}, we employed interpolation operators and multivariate interpolation and extrapolation operators to explore Stable Diffusion’s perception of concepts and spatial information. These operators, based on interpolation estimation methods, leverage multiple embedded real data points as inputs. The individual and collective semantic information of these inputs contribute valuable properties to the creative process. However, interpolation-based operators are not the only options; we will discuss alternative operators in the Discussion and Future Work sections.

\subsection{Query-wise Concept Latent Operation}
The Query-wise Concept Latent Operation focuses on manipulating embedding vectors within the conceptual latent space. The concept vector operator $\mathcal{T}_c$ acts during the denoising process of the network, specifically in each cross-attention block of the U-Net. It operates on the resulting vectors of the attention queries. \cref{fig:pipeline}(b) demonstrate the detail of this process inside a single attention block. The attention queried vectors represent the degree of alignment between the partially generated image and the input prompt in the latent space. This dynamic representation captures the neural network’s evolving understanding of the target concept in conjunction with the image being generated, rather than relying on static memory. Therefore, \workname strategically applies latent space operations at this stage to influence the network's conceptual cognition.

\subsection{Conditioning Vector Shape Latent Operation}
The Conditioning Vector Shape Latent Operation focuses on manipulating shape embedding vectors. The shape vector operator $\mathcal{T}_s$ is applied to the vectors injected into the U-Net at each ControlNet block, as \cref{fig:pipeline}(c) visualize. ControlNet inherently produces vectors with desirable properties because, during its training, the U-Net weights are kept frozen, and ControlNet is trained to compute a bias for each U-Net layer’s output. This bias provides additional control and constraints to the generated results.

Biases from the same input can be combined through simple addition to jointly influence the generation process. Similarly, biases derived from different control images can be subjected to various vector operations, producing meaningful control vectors that enrich the manipulation of generated shapes and spatial attributes.
\section{Case Study}
Based on \workname, we created two distinct artworks: \textit{Infinitepedia} and \textit{Latent Motion}. These two pieces each focus on exploring the potential of AI’s understanding in different domains, namely concepts and shapes, by utilizing the respective operators. Infinitepedia relies on the \textbf{Query-wise Concept Latent Operation}, which manipulates the latent space associated with conceptual vectors. On the other hand, Latent Motion leverages the \textbf{Conditioning Vector Shape Latent Operation}, which controls AI's spatial and shape perception through the manipulation of latent vectors related to motion and form. 

\subsection{Infinitepedia}
\begin{figure}
    \centering
    \includegraphics[width=0.95\linewidth]{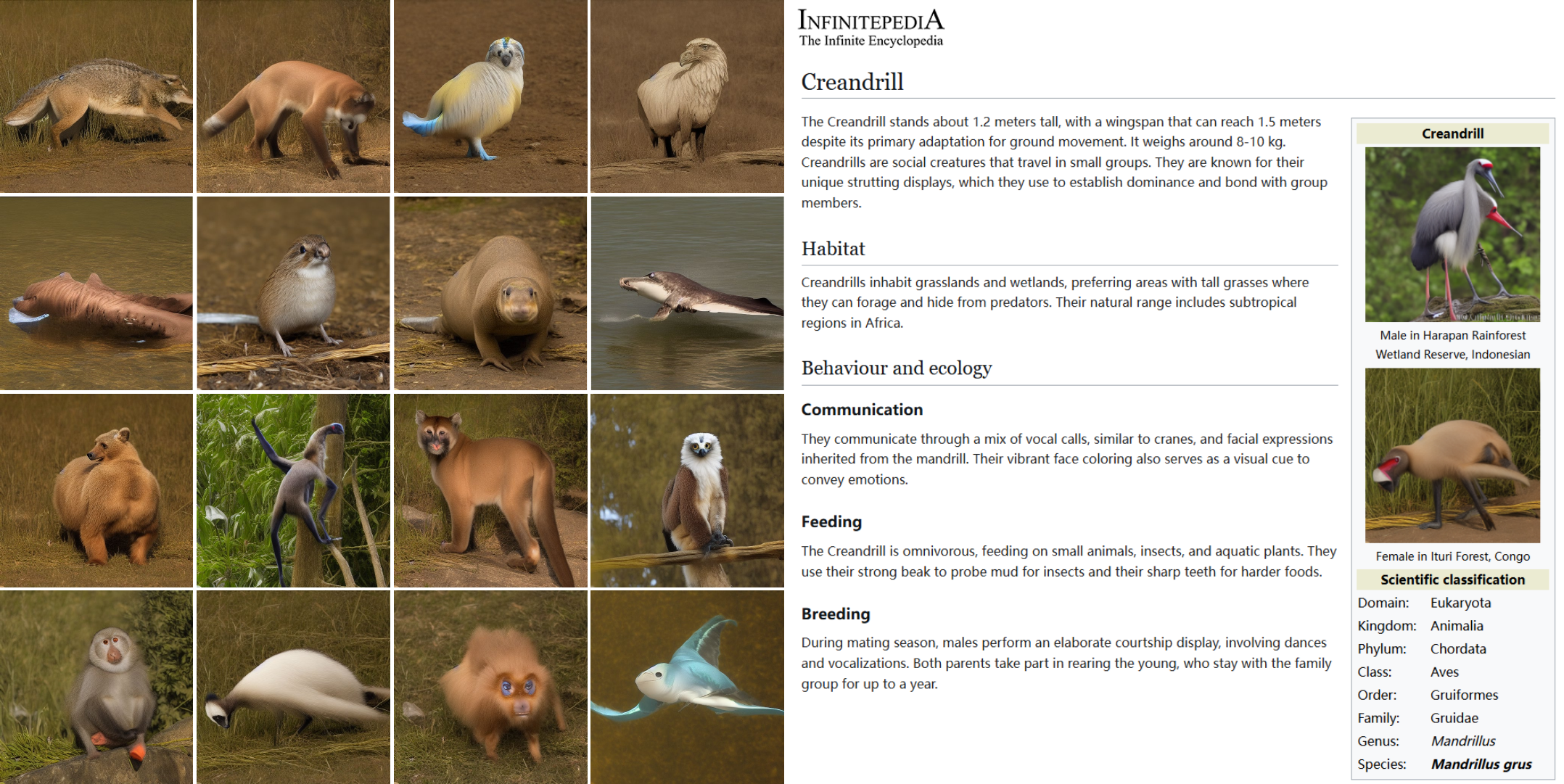}
    \caption{Page and creatures generated in Infinitepedia.}
    \label{fig:pedia}
\end{figure}
\textit{Infinitepedia} is an autonomous generative system that creates an endless encyclopedia of fictional creatures, blending characteristics of real-world species through AI-driven synthesis. Using \workname, \textit{Infinitepedia} manipulates the latent space by embedding two distinct species' conceptual vectors and interpolating them in each attention query through operator $\mathcal{T}_c$ to create a hybrid representation. This process captures Stable Diffusion’s dynamic conceptual understanding and transforms it into visual outputs that merge reality and imagination.

For instance, the system generates an image of a hybrid creature, such as a fusion of a pelican and a panda, and complements it with a textual description generated by ChatGPT. 
During this process, the AI projects its understanding of the words "pelican" and "panda" into corresponding latent space vectors. These vectors represent Stable Diffusion's conceptual comprehension of the two terms. As the denoising process passes through the attention blocks, the latent space representation of the generating image is compared against these two conceptual queries respectively. By interpolating the results of these queries, the system generates a fused image that integrates elements of both concepts.
This artistic exploration showcases the potential of \textbf{Query-wise Concept Latent Operations}, as artists can intervene at the conceptual level, guiding AI's perception of hybridized identities.

\textit{Infinitepedia} exemplifies how conceptual latent space manipulation can inspire new artistic directions, providing tools for artists to explore AI’s interpretive processes and generate imaginative outputs that extend beyond traditional biological representations.
\subsection{Latent Motion}
\begin{figure}
    \centering    \includegraphics[width=0.95\linewidth]{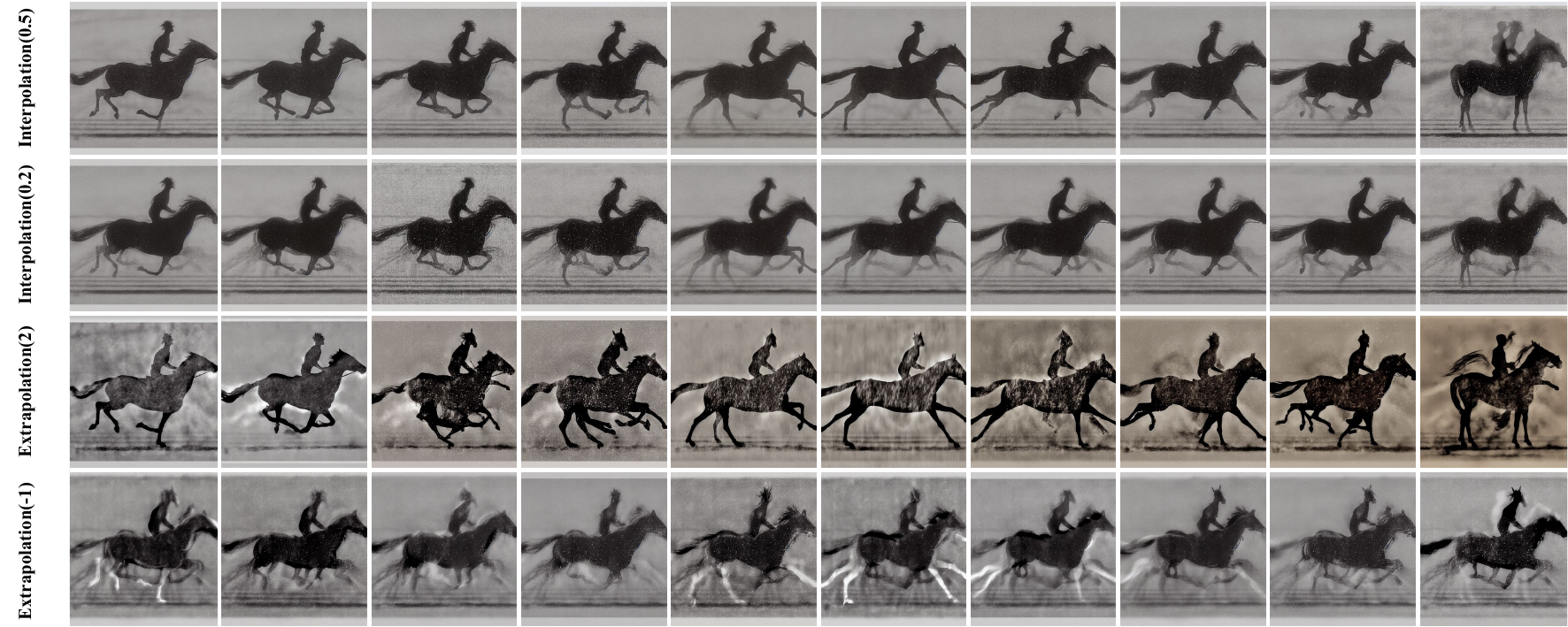}
    \caption{Representative images in Latent Motion.}
    \label{fig:pediapage}
\end{figure}
\textit{Latent Motion} originally started as an experimental component during the development of \workname, where we explored how AI could represent motion and temporal transitions in latent space. It was a critical step in understanding the potential of latent space manipulation for motion representation. However, as the experiment progressed and the results began to reveal AI’s unique interpretations of motion and space, \textit{Latent Motion} evolved into a standalone artistic piece. Inspired by Eadweard Muybridge’s sequential photography\cite{Prodger2003} and Marcel Duchamp’s abstract depiction of motion\cite{duchamp1912nude}, this project uses operator $\mathcal{T}_s$ to explore the latent space of shape and spatial perception.

The experiment embeds multiple sequential motion frames from Muybridge’s The Horse in Motion into the latent space, constructing a high-dimensional convex hull. Using multivariate interpolation and extrapolation techniques, this convex hull serves as a framework for visualizing AI's understanding of motion.

This approach reveals how the neural network perceives and synthesizes motion, challenging traditional notions of temporal continuity. Through interpolation, AI blends motion states into smooth transitions, while extrapolation pushes beyond the original frames to create novel, abstract representations. \textit{Latent Motion} exemplifies the potential of \textbf{Conditioning Vector Shape Operations} to redefine AI-driven representations of motion, offering an artistic lens on spatial manipulation within latent space.

\section{Results and Discussion}
\subsection{Concept Latent Operation}
In the \textbf{Query-wise Concept Latent Operation}, we use the query results as the operation position for the concept operator. An alternative design is to use the embedded vector projected from the prompts as the operation position for the concept operator, which we refer to as feature-wise. Through a series of experiments, we found that query-wise exhibits superior properties when performing various operations. The results of these operations are more accurately representative of the concepts described by the prompt and are less likely to fall into "meaningless" areas within latent space. This makes query-wise a more effective approach in maintaining the integrity of the conceptual representation.
\begin{figure}
    \centering
    \subfloat[]{\includegraphics[width=0.95\linewidth]{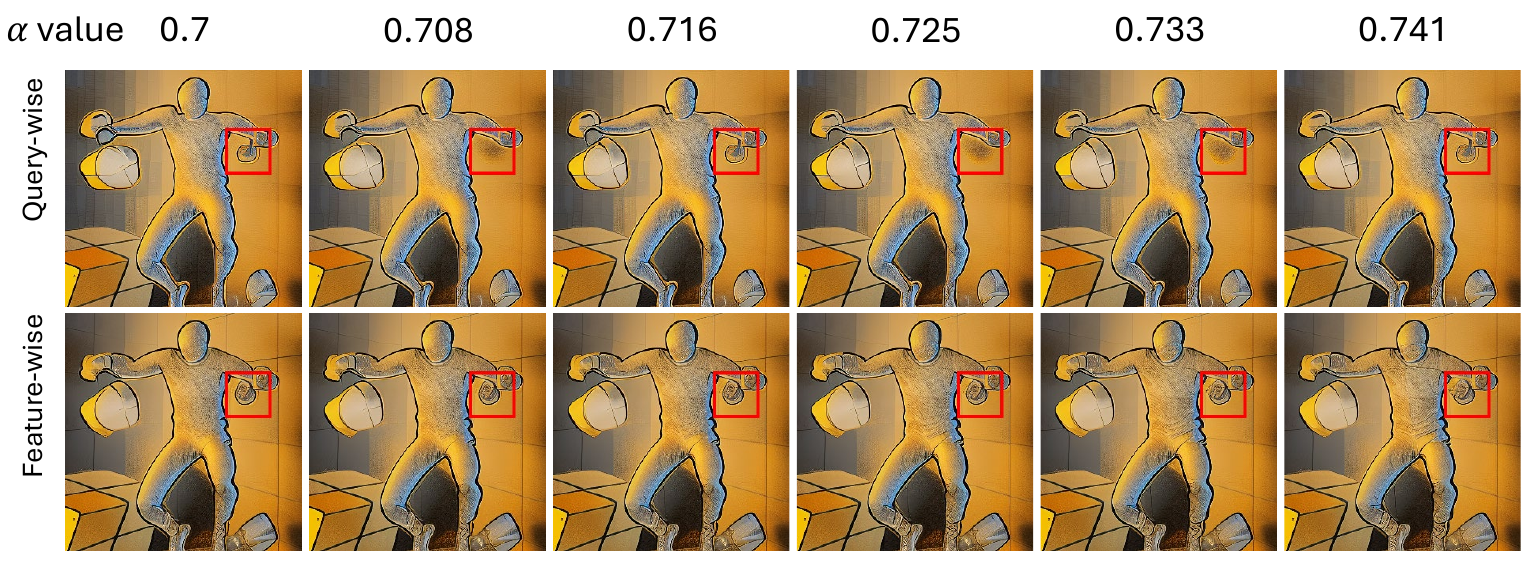}}
    \\
    \subfloat[] {\includegraphics[width=0.95\linewidth]{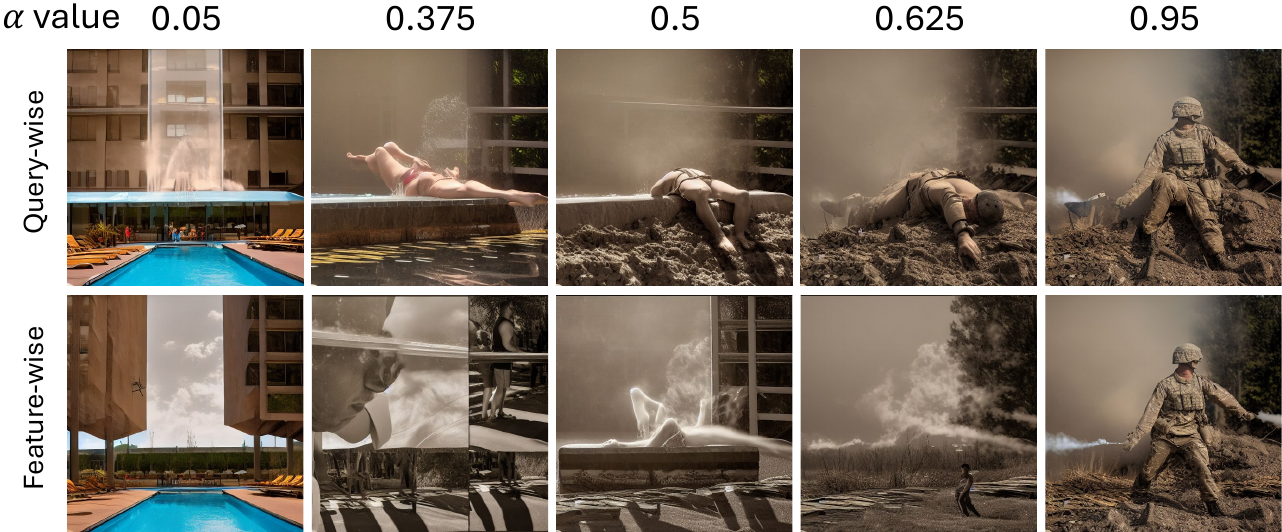}}  
    \caption{Results with(a) and without(b) ControlNet. Left side prompt is "A man swimming in the pool." While right side prompt is "A soldier blown into the air by a fragment grenade."}
    
    \label{fig:result}
\end{figure}
To compare these two approaches, we designed a series of experiments where we interpolated between two sets of prompts. We tested both with and without ControlNet, and found that at the endpoints of the interpolation, there was no significant difference between the two methods. However, at an interpolation alpha value near 0.5, the point furthest from the each endpoints, the two methods produced notably different output results.

\cref{fig:result} presents representative samples from the experimental data. \cref{fig:result}(a) shows the results without ControlNet. When the alpha value is close to 0 or 1, where either the pool prompt or the battlefield prompt dominates, the outputs from both methods converge. However, when interpolated towards the middle, the query-wise approach effectively generates a blend of both prompts, while the feature-wise approach leads to confusion (as seen in the case of alpha = 0.375) or produces results that are less relevant to either of the input prompts (as in the case of alpha = 0.625).

\cref{fig:result}(b) shows a series of interpolation results when ControlNet is used. In the feature-wise results, flickering appears in the highlighted areas (red box). With ControlNet, the network's output is controlled by both the prompt and the control input. The flickering indicates that the latent space vectors mixed using feature-wise are ambiguous or meaningless to the AI. The control input with clear and practical meaning, combined with vectors resulting from operations that the network cannot properly decode, ends up conflicting with each other, leading to the flickering effect. Therefore, we conclude that the query-wise approach is a better choice for conceptual latent space manipulation.
\subsection{Latent Field}
In latent space, there exists a "field" where information is unevenly distributed, with "meaningful volumes," "semantically ambiguous volumes," and "meaningless volumes." 
This "field" is implicitly defined by the subsequent neural network. The meaning of a latent space vector is not intrinsic but is derived through the network’s interpretation, which ultimately transforms the vector into a human-understandable form. The information contained in these vectors only acquires meaning once it is processed by the network.
\begin{figure}
    \centering
    \subfloat{\includegraphics[width=0.20\linewidth]{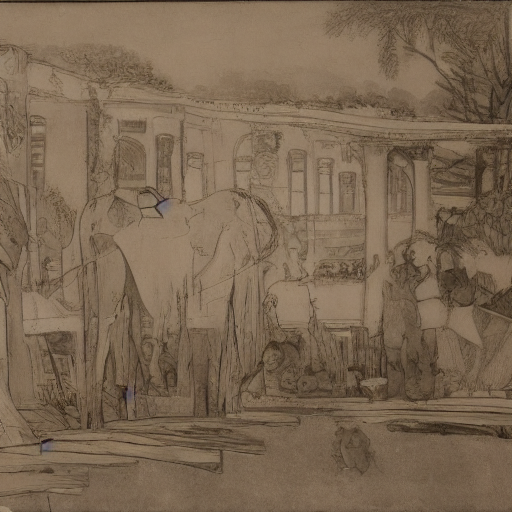}}
    \hspace{4pt}
    \subfloat{\includegraphics[width=0.20\linewidth]{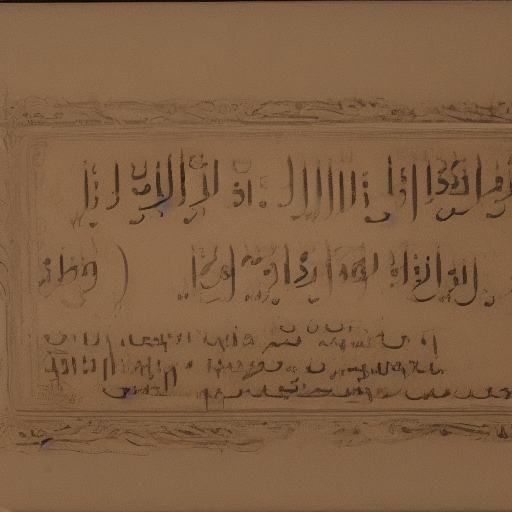}} 
    \hspace{4pt}
    \subfloat{\includegraphics[width=0.20\linewidth]{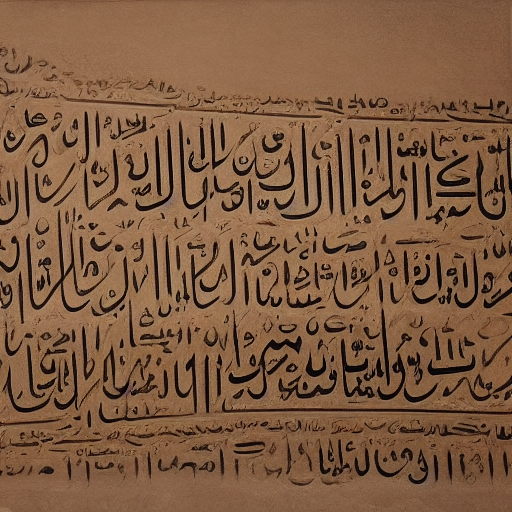}}
    \hspace{4pt}
    \subfloat{\includegraphics[width=0.20\linewidth]{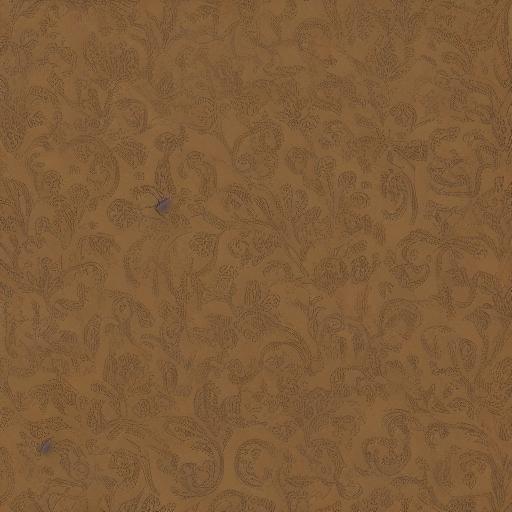}} 
    \caption{Images in the "deserts" of latent space. Occurs when the latent space vectors contain information that the AI cannot interpret, such as when the prompt consists of pure numbers, partial non-English text, or when vector manipulations deliberately push the latent space vector far from the volumes that have meaning to neural network.}
    
    \label{fig:desert}
\end{figure}
During our experiments, we explored many meaningless volumes in latent space. These volumes resemble "deserts" within latent space. Whenever the exploration falls into these volumes, Stable Diffusion tends to produce outputs resembling drawings on parchment paper or inscriptions in Arabic carved into stone, as shown in \cref{fig:desert}. The results from the feature-wise experiments validate this observation. During the feature-wise exploration, the latent space often passed through semantically ambiguous volumes. If we project the linear interpolation path of query-wise into the feature-wise latent space, it might trace a high-dimensional curve that exclusively passes through meaningful volumes.

Past explorations of latent space typically focused on analyzing neighborhoods around specific data points or the vectors they form. In future work, we plan to explore the overall structure of latent space, attempting to identify, color-code and visualize distinct regions within this space, creating a more comprehensive map of its features. This approach could help us better understand the latent space, making the field of meaningful and meaningless volumes more tangible and visually interpretable.

\subsection{Operation in Latent Space}
Latent space can also be understood as a high-dimensional space. While the two case study artworks discussed in this paper primarily explore latent space through interpolation-based operations, other forms of spatial geometry can also be applied to latent space vectors. In our exploration within Latent Motion, we already began to shape the idea of constructing a framework within latent space using existing data. Moving forward, we plan to experiment with more complex operations on latent space vectors. These may include projecting new data points into the established framework, or constructing new orthogonal vectors through the cross-product of existing vectors. By applying such advanced geometric operations, we aim to unlock further creative potential within latent space manipulation, expanding the range of artistic and computational possibilities in the future work.

\section{Conclusion}
This paper presented Latent Diffusion, a framework designed to enhance Stable Diffusion by incorporating customizable latent space operations for conceptual and spatial manipulation. By allowing artists to intervene directly in the generative process, this tool provides new flexibility and precision for exploring latent representations. The case studies Infinitepedia and Latent Motion demonstrated practical applications, showcasing the potential for hybrid concept generation and spatial transformations. Additionally, our exploration of latent regions identified structural properties such as meaningful trajectories and ambiguous areas, suggesting new avenues for geometric approaches to latent space analysis. Latent Diffusion contributes a dynamic layer of interpretability and control to diffusion models, bridging technical and artistic innovation in AI-driven generative art.

\bibliographystyle{ACM-Reference-Format}
\bibliography{main}










\end{document}